\documentclass{article}

    \usepackage[nonatbib,preprint]{neurips_2020}

\usepackage[utf8]{inputenc} 
\usepackage[T1]{fontenc}    
\usepackage{hyperref}       
\usepackage{url}            
\usepackage{booktabs}       
\usepackage{amsfonts}       
\usepackage{nicefrac}       
\usepackage{microtype}      

\usepackage{url}
\usepackage{algorithm}
\usepackage{algorithmic}
\usepackage{graphicx}
\usepackage{amsthm}
\usepackage{amssymb}
\usepackage{multicol}
\usepackage{amsmath}
\usepackage{wrapfig}

\usepackage{scrwfile}
\TOCclone[\contentsname~(\appendixname)]{toc}{atoc}

\AfterTOCHead[toc]{%
}
\AfterTOCHead[atoc]{%
  \edef\maintocdepth{\the\value{tocdepth}}%
  \value{tocdepth}=-10000\relax%
}

\title{Online Semi-Supervised Learning in \\Contextual Bandits with Episodic Reward}

\author{%
  Baihan Lin \\
  Columbia University\\
  \texttt{baihan.lin@columbia.edu} \\
}

\begin{document}

\maketitle

\begin{abstract}

We considered a novel practical problem of online learning with episodically revealed rewards, motivated by several real-world applications, where the contexts are nonstationary over different episodes and the reward feedbacks are not always available to the decision making agents. For this online semi-supervised learning setting, we introduced Background Episodic Reward LinUCB (BerlinUCB), a solution that easily incorporates clustering as a self-supervision module to provide useful side information when rewards are not observed. Our experiments on a variety of datasets, both in stationary and nonstationary environments of six different scenarios, demonstrated clear advantages of the proposed approach over the standard contextual bandit. Lastly, we introduced a relevant real-life example where this problem setting is especially useful. 
\footnote{The data and codes can be accessed at \href{https://github.com/doerlbh/BerlinUCB}{\underline{https://github.com/doerlbh/BerlinUCB}}}

\end{abstract}
\section{Introduction}

Online learning is a common problem in many practical applications where the data become available in a sequential order and later used to update the best predictor for future data or reward associated with the data features. In many case, the reward feedback is the only source where the online learning agent can effectively learn from the sequential past experience. This problem is especially important in the field of sequential decision making where the agent must choose the best possible action to perform at each step to maximize the cumulative reward over time. One key challenge is to obtain an optimal trade-off between the exploration of new actions and the exploitation of the possible reward mapping from known actions. This framework is usually formulated as the \textit{Multi-Armed Bandits (MAB)} problem where each arm of the bandit corresponds to an unknown (but usually fixed) reward probability distribution \cite{LR85,UCB}, and the agent selects an arm to play at each round, receives a reward feedback and updates accordingly.

However, in many real-world applications, the reward feedback are not always revealed. For instance, in a recommendation system \cite{MaryGP15}, the users might not be attending to and interacting with the system all the time, and their inactivity should be considered a latent mixture of positive and negative feedbacks, instead of merely a negative feedback in traditional online learning frameworks. In personalized medicine or longitudinal studies \cite{villar2015multi}, the patients might not conduct follow up check ups for many reasons, and while these data points are missing, their contexts (demographics, genders, ages etc.) might still be useful to predict medical outcomes for other people (e.g. to gain a better understanding of the  patients population). In many interactive systems, the goal is to provide an efficient assistance without unnecessary human intervention, and in these cases, the agent would need to learn from empty feedback in most of the time. In addition, the reward feedback can come in different episodes or frequencies. For instance, an advertisement website might experience a surge in clicking in the golden hours, while not so much in off seasons. An effective system should adapt to the nonstationarity in these observed rewards \cite{lin2020diabolic}. To study these scenarios, we proposed a novel problem setting called \textit{Online Learning with Episodic Reward}.

In this paper, we study this novel problem setting with the \textit{Contextual Bandits (CB)}, a useful variant of MAB where at each step, the agent observes an $N$-dimensional \textit{context}, or \textit{feature} vector before selecting an action. Theoretically, the ultimate goal of CB is to learn the relationship between the rewards and the context vectors so as to make better decisions given the context \cite{AgrawalG13}. The idea is that whenever the reward is missing, an agent can still have access to a population of labelled and unlabeled context history (i.e., contexts without the associated rewards), which can be potentially used as a prior knowledge to improve the subsequent online decision-making. We adopted a self-supervised learning techniques, which is a form of unsupervised learning where the data population provides the supervision. By installing a clustering module, in the online mode without associate reward feedback, we can decide which arm-related representations to update for a given context; such content-driven representation selection has a potential to further improve the subsequent decision-making. These flexible self-supervision modules can (and should) continue to be updated online as more contexts become available, especially in {\em nonstationary environments} abundant in practical applications, where both the context distribution and the reward distribution can change in various ways.
 
We evaluate our approach, \textit{Background Episodically Rewarded LinUCB (BerlinUCB)}, on several types of nonstationary environments and demonstrated that (1) updating the representation structure when no reward is revealed, in general, considerably improves performance of contextual bandit; and (2) moreover, in many cases, adaptive learning with context-dependent clustering modules is much better than learning without the self-supervision. 



 

\section{Related Work}
\label{sec:related}

There have been multiple solutions proposed for sequential decision making and the contextual bandit problems \cite{lin2020online,lin2020unified,lin2020astory,lin2019split}. LinUCB assumes a linear dependency between the expected reward of an action and its context \cite{Li2010,ChuLRS11}. Its representation space is modeled using a set of linear predictors. This assumption is not used in Neural Bandit \cite{AllesiardoFB14}. However, these algorithms assume that the agent can observe the reward at each iteration, which is not the case in many practical applications as discussed earlier. 

To consider the problem of incomplete feedbacks, \cite{bartok2014partial} studies ``Partial  Monitoring (PM)'', which allows the learner to retrieve the expected value of actions through an analysis of the feedback matrix. In our work, we don't hold the same assumption that this feedback matrix is always known to the learner. To consider the corrupted rewards, \cite{gajanecorrupt} studies a variant of the stochastic bandit where the goal is to maximize the sum of the unobserved rewards given the observation of transformation of these rewards through a stochastic corruption process. Unlike their approach which relies on known parameters for those transformations, we directly build upon LinUCB with no additional parameters. Another related work is to introduce different representation modules such as autoencoders that adaptively update across batches \cite{lin2018contextual}. However, it only uses the unlabelled data for pretraining, and during the online mode, the reward is always revealed.


\section{Problem setting}

\subsection{The Contextual Bandit problem} 
Following \cite{langford2008epoch}, the contextual bandit problem is defined as follows.
At each time point (iteration) $t \in \{1,...,T\}$, an agent is presented with a {\em context} ({\em feature vector}) $\textbf{x}_t \in \mathbf{R}^N$
  before choosing an arm $k  \in A = \{ 1,...,K\} $.
We will denote by
  $X=\{X_1,...,X_N\}$  the set of features (variables) defining the context.
Let ${\textbf r_t} = (r^{1}_t,...,$ $r^{K}_t)$ denote  a reward vector, where $r^k_t \in [0,1]$ is a reward at time $t$  associated with the arm $k\in A$.
Herein, we will primarily focus on the Bernoulli bandit with binary reward, i.e. $r^k_t \in \{0,1\}$.
Let $\pi: X \rightarrow A$ denote a policy.  Also, $D_{c,r}$ denotes a joint distribution over  $({\bf x},{\bf r})$.
We will assume that the expected reward is a  linear function of the context, i.e.
$E[r^k_t|\textbf{x}_t] $ $= \mu_k^T \textbf{x}_t$,
where $\mu_k$ is an unknown weight vector (to be learned from the data) associated with arm $k$. 

\subsection{Online Semi-Supervised Learning with Episodic Reward}
Algorithm \ref{alg:episode} presents at a high-level our problem setting, where the context $\textbf{x}(t)\in \mathbb{R}^d$ is a vector describing the context $C$ at time $t$, $r_{a_t,t}(t) \in [0,1]$ is the reward of action $a_t$ at time $t$, and $\textbf{r}(t) \in [0,1]^K$ denotes a vector of rewards for all arms at time $t$. $\mathbb{P}_{x,r}$ denotes a joint probability distribution over $(x,r)$, and $\pi: C \rightarrow A$ denotes a policy. Unlike traditional setting, in step 5 we have the rewards revealed in an episodic fashion (i.e. sometimes there are feedbacks of rewards being 0 or 1, sometimes there are no feedbacks of any kind), given by a probability $p_r \in [0,1]$. We consider our setting an online semi-supervised learning problem \cite{Yver2009,ororbia2015online}, where the agent learns from both labeled and unlabeled data in online setting. However, in their setting the true label is received at each iteration, while in our setting a bandit feedback is assumed, i.e., if classification was incorrect, the agent will not know what the correct label was, only that its decision was incorrect.  

\begin{algorithm}[tb]
 \caption{Online Learning with Episodic Reward}
 \label{alg:episode}
 \begin{algorithmic}[1]
 \STATE {\bfseries }\textbf{for} t = 1,2,3,$\cdots$, T \textbf{do}
\STATE {\bfseries } \quad $(\textbf{x}(t),\textbf{r}(t))$ is drawn according to $\mathbb{P}_{x,r}$
\STATE {\bfseries }\quad Context $\textbf{x}(t)$ is revealed to the player
\STATE {\bfseries }\quad  Player chooses an action $a_t =\pi_t(\textbf{x}(t))$
\STATE {\bfseries } \quad Feedback $r_{a_t,t}(t)$ for only chosen arms are episodically revealed
\STATE {\bfseries } \quad Player updates its policy $\pi_t$
\STATE {\bfseries } \textbf{end for}
 \end{algorithmic}
\end{algorithm}

\section{Background Episodically Rewarded LinUCB}

We proposed Background Episodically Rewarded LinUCB (BerlinUCB), a semi-supervised and self-supervised online contextual bandit which updates the context representations and reward mapping separately given the state of the feedbacks being present or missing (Algorithm \ref{alg:berlinucb}). We assume that (1) when there are feedbacks available, the feedbacks are genuine, assigned by the oracle, and (2) when the feedbacks are missing (not revealed by the background), it is either due to the fact that the action is preferred (no intervention required by the oracle, i.e. with an implied default rewards), or that the oracle didn't have a chance to respond or intervene (i.e. with unknown rewards). Especially in the Step 15, when there is no feedbacks, we assign the context $\textbf{x}_t$ to a class $a'$ (an action arm) with the self-supervision given the previous labelled context history. Since we don't have the actual label for this context, we would only update the reward mapping parameter $\textbf{b}_{a'}$ and leave the covariance matrix $\textbf{A}_{a'}$ untouched. This additional usage of unlabelled data (or unrevealed feedback) is especially important in our model, as it allows the agent to learn even when the reward is not available. 

\begin{algorithm}[tb]
 \caption{BerlinUCB}
 \label{alg:berlinucb}
 \begin{algorithmic}[1]
 \STATE {\bfseries } \textbf{Initialize} $c_t \in \mathbb{R}_+, \textbf{A}_a \leftarrow \textbf{I}_d, \textbf{b}_a \leftarrow \textbf{0}_{d \times 1} \forall a \in \mathcal{A}_t $
 \STATE {\bfseries }\textbf{for} t = 1,2,3,$\cdots$, T \textbf{do}
  \STATE {\bfseries } \quad Observe features $\textbf{x}_{t}\in\mathbb{R}^d$
 \STATE {\bfseries } \quad \textbf{for all} $a \in \mathcal{A}_t $ \textbf{do}
 \STATE {\bfseries }  \quad \quad  $\hat{\mathbf{\theta}}_a \leftarrow \textbf{A}_a^{-1}\textbf{b}_a$
 \STATE {\bfseries } \quad \quad $p_{t,a} \leftarrow \hat{\mathbf{\theta}}_a^{\top}\textbf{x}_{t}+c_t\sqrt{\textbf{x}_{t}^{\top}\textbf{A}_a^{-1}\textbf{x}_{t}}$
\STATE {\bfseries }  \quad \textbf{end for}
\STATE {\bfseries } \quad Choose arm $a_t=\arg\max _{a \in \mathcal{A}_t } p_{t,a}$ 
\STATE {\bfseries } \quad \textbf{if} the background revealed the feedbacks \textbf{then} 
 \STATE {\bfseries }  \quad \quad Observe feedback $r_{a_t,t}$ 
\STATE {\bfseries }  \quad \quad $\textbf{A}_{a_t} \leftarrow \textbf{A}_{a_t}+\textbf{x}_{t}\textbf{x}^{\top}_{t}$ 
\STATE {\bfseries }  \quad \quad $\textbf{b}_{a_t} \leftarrow \textbf{b}_{a_t}+r_{a_t,t}\textbf{x}_{t}$ 
\STATE {\bfseries } \quad \textbf{elif} the background revealed NO feedbacks \textbf{then} 
\STATE {\bfseries } \quad \quad  \textbf{if} use self-supervision feedback
\STATE {\bfseries } \quad \quad  \quad $r' = [a_t == \text{predict}(\textbf{x}_{t})] $ \hfill \% clustering modules
\STATE {\bfseries } \quad \quad \quad  $\textbf{b}_{a_t} \leftarrow \textbf{b}_{a_t}+r'\textbf{x}_{t}$ 
\STATE {\bfseries } \quad \quad \textbf{elif} \hfill \% ignore self-supervision signals
\STATE {\bfseries }  \quad \quad  \quad $\textbf{A}_{a_t} \leftarrow \textbf{A}_{a_t}+\textbf{x}_{t}\textbf{x}^{\top}_{t}$ 
\STATE {\bfseries } \quad \quad \textbf{end if}
\STATE {\bfseries } \quad \textbf{end if}
 \STATE {\bfseries }\textbf{end for}
 \end{algorithmic}
\end{algorithm}

\subsection{The Self-Supervision and Semi-Supervision Modules}

We construct our self-supervision modules given the cluster assumption of the semi-supervision problem: the points within the same cluster are more likely to share a label. As shown in many applications, clustering algorithms like Kmeans, Gaussian Mixture Model (GMM) \cite{reynolds2009gaussian} and spectral clustering \cite{von2007tutorial} are especially powerful unsupervised modules, especially in their offline versions. Their online variants, however, often performs poorly \cite{zhang2019fully}. As in this work, we focus on the completely online setting, we chose three popular clustering algorithms as the self-supervision modules: GMM, Kmeans and K-nearest neighbors.

In the following section, we empirically evaluated the performance of 5 agents:  LinUCB is the standard contextual bandit which treated missing feedbacks as a negative reward (of zero). BerlinUCB is the standard version of our proposed contextual bandit, designed for sparse feedbacks without the self-supervision modules. It updates the context representations but not the reward mapping. Among the BerlinUCB family, we also compared three BerlinUCB variants with the self-supervision clustering modules: Kmeans, KNN (with K=5) and GMM, which we denoted: B-Kmeans, B-KNN, and B-GMM, respectively.

\section{Empirical Evaluation}

We evaluated our approach on six synthetic nonstationary online learning scenarios in two datasets: MNIST \cite{lecun1998mnist} and Warfarin \cite{international2009estimation}, and provided a real-life application where the episodic feedbacks are naturally installed: a speaker diarization task with sporadic human interactions as bandit feedbacks. For the synthetic online learning task, in order to simulate an online data stream, we draw samples from each dataset sequentially, starting from the beginning each time we draw the last sample. At each round, the algorithm receives  reward 1 if the instance is classified correctly, and 0 otherwise. We compute both the final cumulative reward and the total number of classification errors as the performance metrics. However,  Warfarin dataset is different, as it was actually produced in a real bandit setting, rather than classification setting.

\textit{Important distinction on Bandit vs. classification feedback}: It is important to keep in mind that the bandit feedback (correct/incorrect classification) makes the classification problem significantly more challenging,  as compared to the standard supervised learning, since the true label is never revealed in bandit setting unless the classification is correct. Thus, the classification accuracy in a bandit setting is expected to be lower than in the supervised learning setting, which is not due to inferiority of bandit decision making algorithm versus classifiers, but due to increased problem difficulty, i.s. the lack of feedback about what the correct decision should have been. Recall that such bandit feedback is often a much more realistic model of agent's interaction with the world, especially in online decision making applications such as online advertisement, clinical trials, and so on, which do not fit into the classical classification framework.

However, for empirical evaluation purposes, it is common to use available classification datasets to simulate an online environment with the bandit feedback  (i.e., simulating the situation where the bandit receives, for example, 1 or 0 for correct or incorrect decision, but is not told what the correct decision should have been when it receives 0; such feedback is different from standard online classification feedback in case of non-binary classification). We use the classification dataset here to create online learning simulation environments.

We now describe some details of the experiments. For MNIST, we took the 5,000 samples from the training dataset to simulate the online bandit with 10 arms corresponding to different digits. For Warfarin dataset, we selected the first 5,000 training samples to simulate the online bandit with 3 arms (classes). As an even more challenging task, we also created a multi-task scenario where a mixture of the MNIST and the Warfarin task are fed in batches.  We extended the set of possible labels (arms) to include 10 labels from MNIST and 3 labels from Warfarin, into a total of 13 arms. We used linear stretching to make the input dimensions equal across the two domains. In another word, for each input, we concatenate the Warfarin features (of size 93) on the back of the MNIST features (of size 784) into a sparse vector of size 877. The algorithm had to assign a label to each input without any information about which domain the input came from.

\subsection{Nonstationary Environments}

We simulated six types of nonstationarity using the above datasets (with a batch size of 100). As mentioned before, we assume that the input data arrive in batches, and the data distribution (i.e., the joint distribution of the context and reward) may change across those batches, while remaining stationary within each batch. 

\textbf{Nonstationary context: varying cluster distribution.} 
To simulate changes in the context (input) distribution, we first   clustered all samples in the corresponding  pre-training data subset   into $k$ clusters. Next, we generate a sequence of batches, where each  batch   contained a certain fraction of samples from different clusters, and these fractions were changing across the batches, i.e. the probability distribution of cluster membership was changing, simulating nonstationary input.


\textbf{Nonstationary context: negative images.}
Another type of input  nonstationarity involved introducing negative images as inputs with same semantics but different textures. Namely, with probability $p$, the negative image of the original image was  presented as an input. 
Experiments were performed with this data corruption given by $0 < p < 1$ episodically and randomly assigned for each batch.


\textbf{Nonstationary reward: shuffled class labels.}
 We further explored the multi-task setting by introducing a different type of nonstationary reward, where the class labels were shuffled, i.e. randomly permuted, in each batch.


\textbf{Nonstationary reward: multi-task environment.} Here we assume that input samples come from different domains (tasks), and thus can be associated with different subsets of labels (arms). For example, we combined 5,000 randomly selected training samples from each of the two selected domains, MNIST and Warfarin datasets, and extended the set of possible labels (arms) to include 10 labels from MNIST and 3 labels from Warfarin into a mixed dataset of 10,000 samples of 13 arms. We used linear stretching to make the  input dimensions equal across the two domains. The algorithm had to assign a label to each input without any information about which domain the input came from.


\textbf{Nonstationary oracle: fixed vs. extendable arms.} Traditional online learning have a fixed number of arms (denoted ``F''). We here considered such a nonstationarity where we start with a single arm, and when new labels arrive new arms are generated accordingly, a problem loosely modelled by the bandits with infinitely many arms \cite{berry1997bandit}. For this setting, we applied an arm expansion process  (denoted ``E''): starting from a single arm (for adding a ``new'' arm), if a feedback confirms a new addition, a new arm is appended to the arm list.


\textbf{Nonstationary oracle: varying episodic rewards.}
Last but not least, we may assume the probability of the reward revealing ($p_r$ in Algorithm \ref{alg:episode}) can change from batch to batch. In another word, the reward feedback can be more frequently revealed in some episode or mini-batch than the others.

\subsection{Results}

\begin{figure}[tb]
\centering
    \includegraphics[width=0.24\linewidth]{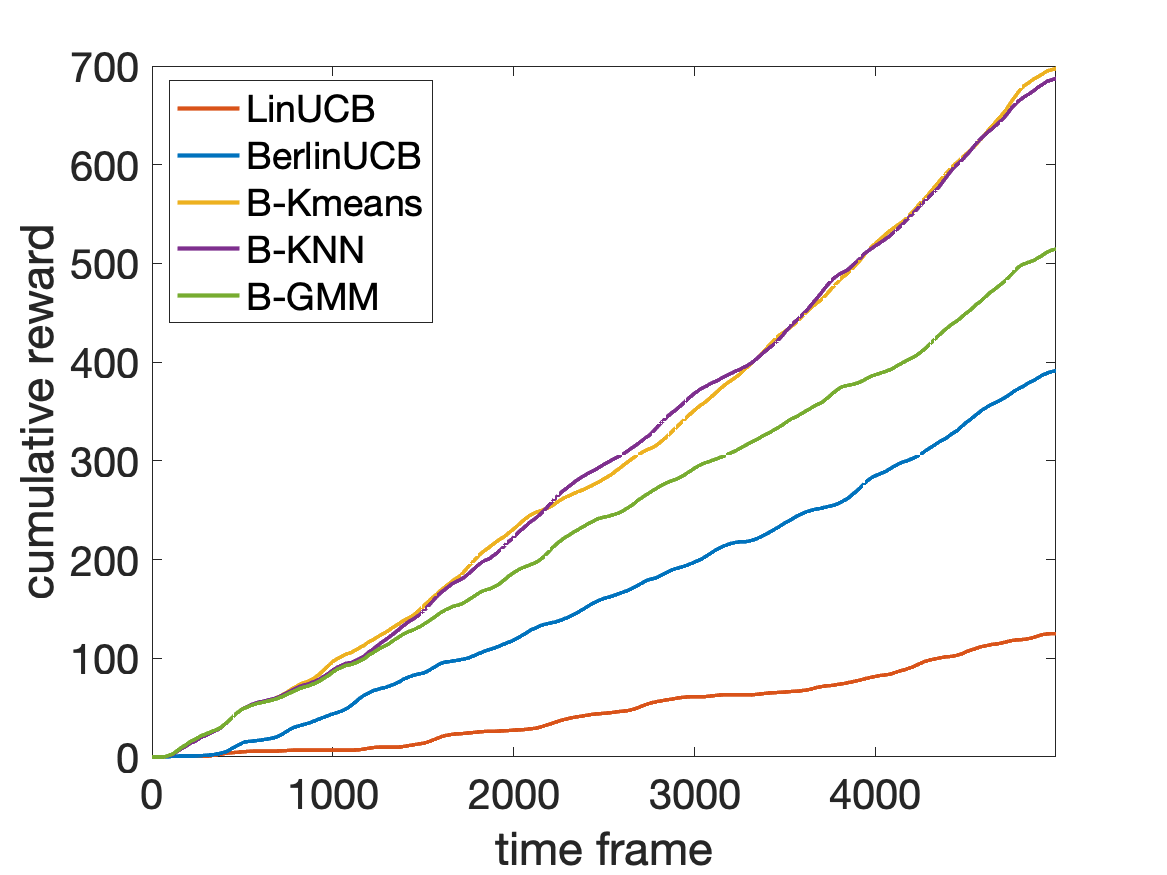}
    \hfill
    \includegraphics[width=0.24\linewidth]{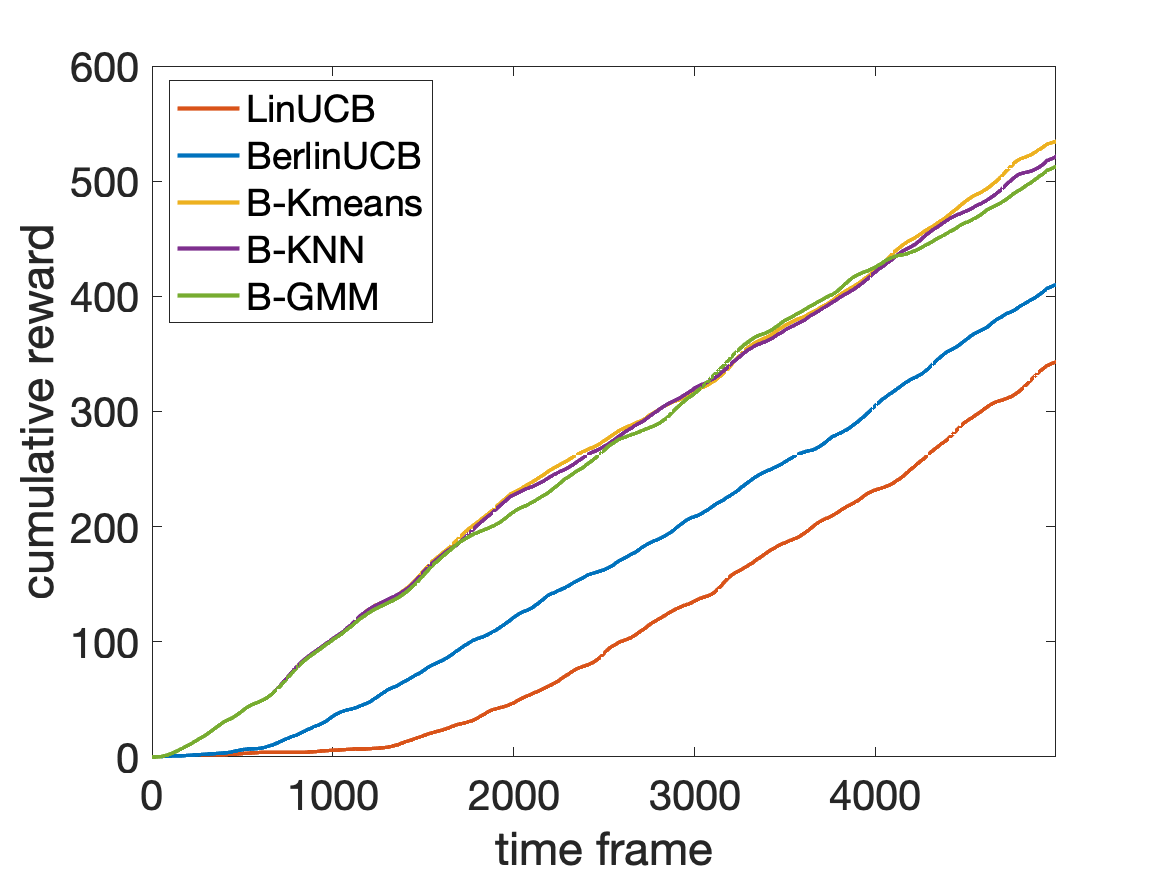} \hfill
    \includegraphics[width=0.24\linewidth]{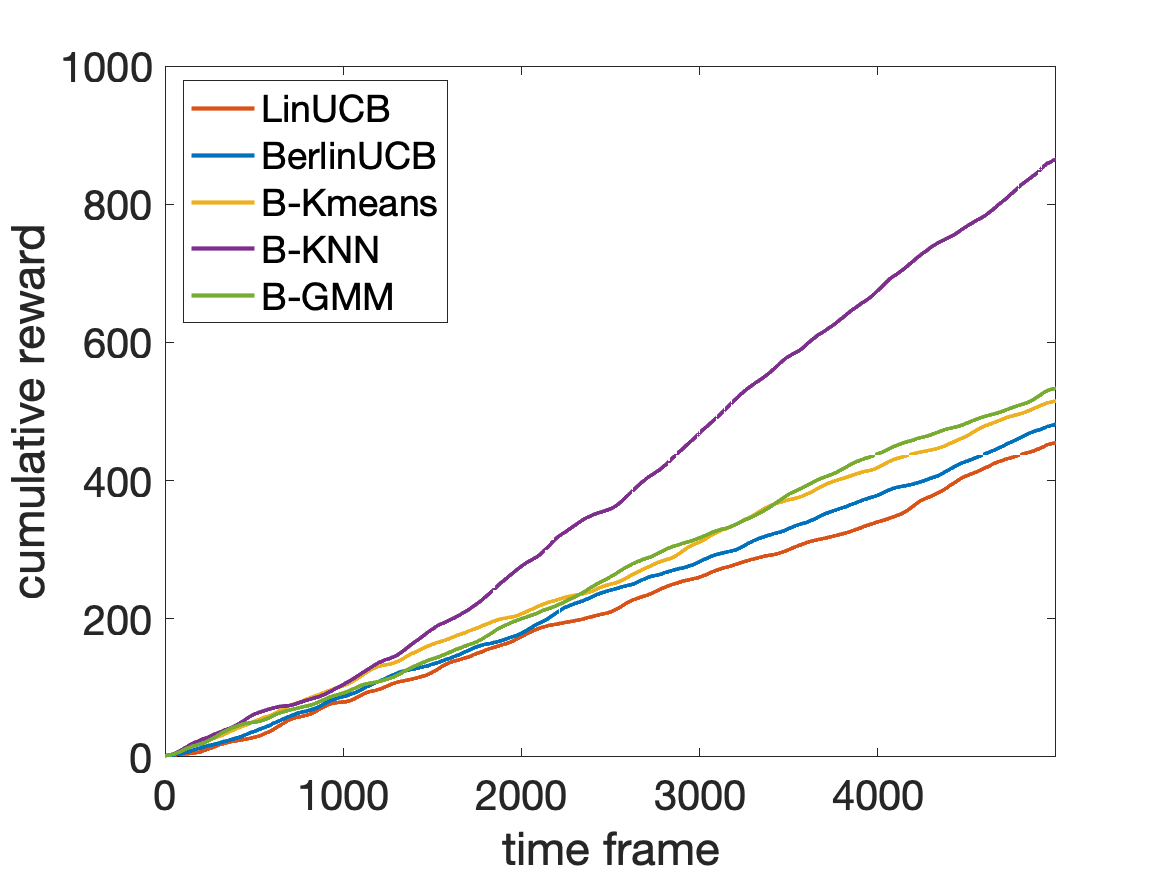} \hfill
    \includegraphics[width=0.24\linewidth]{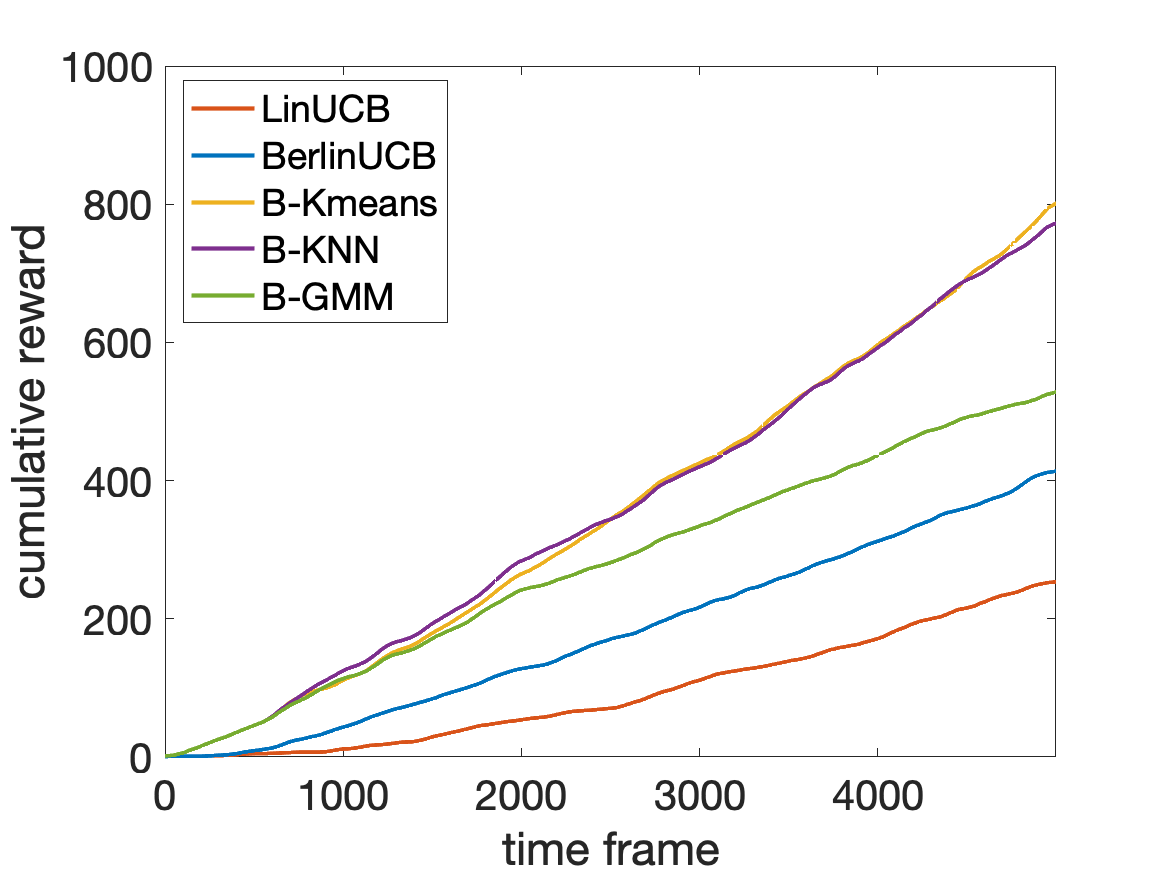}
    \includegraphics[width=0.24\linewidth]{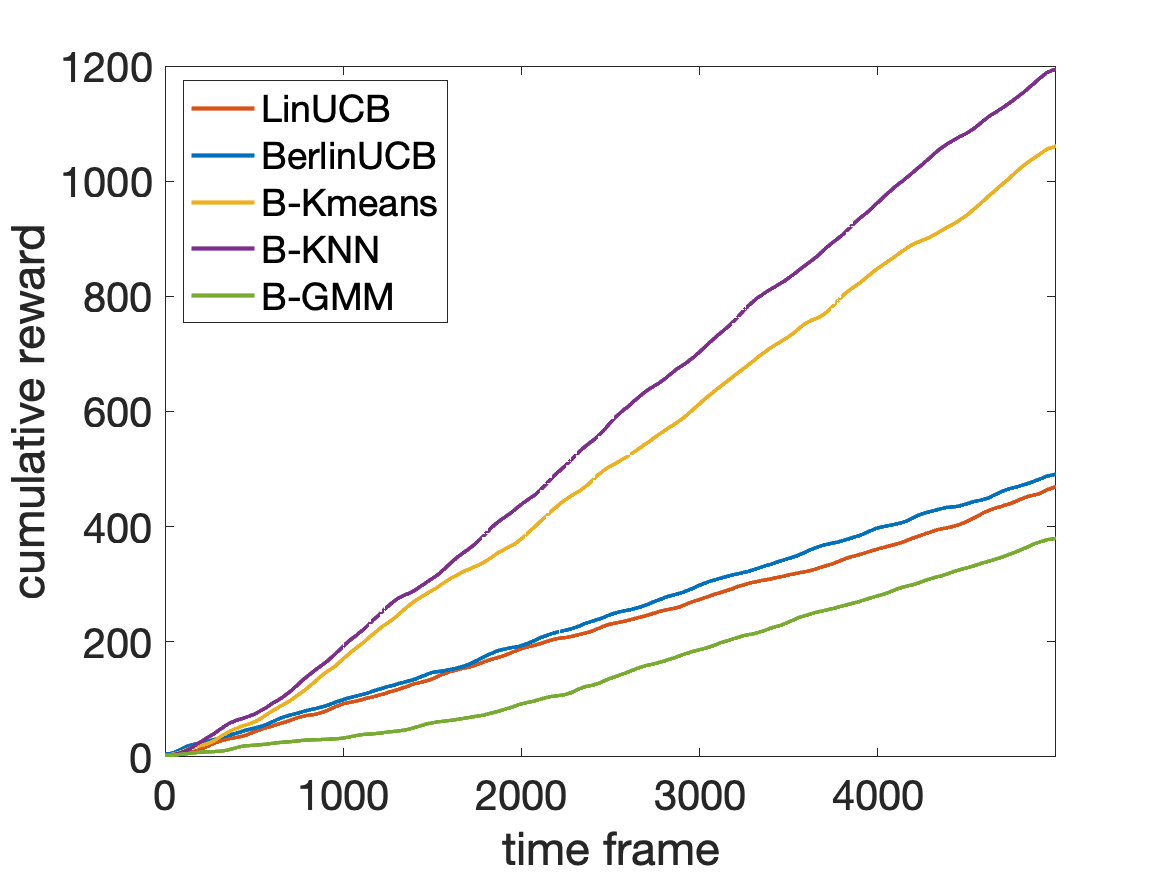} \hfill
\includegraphics[width=0.24\linewidth]{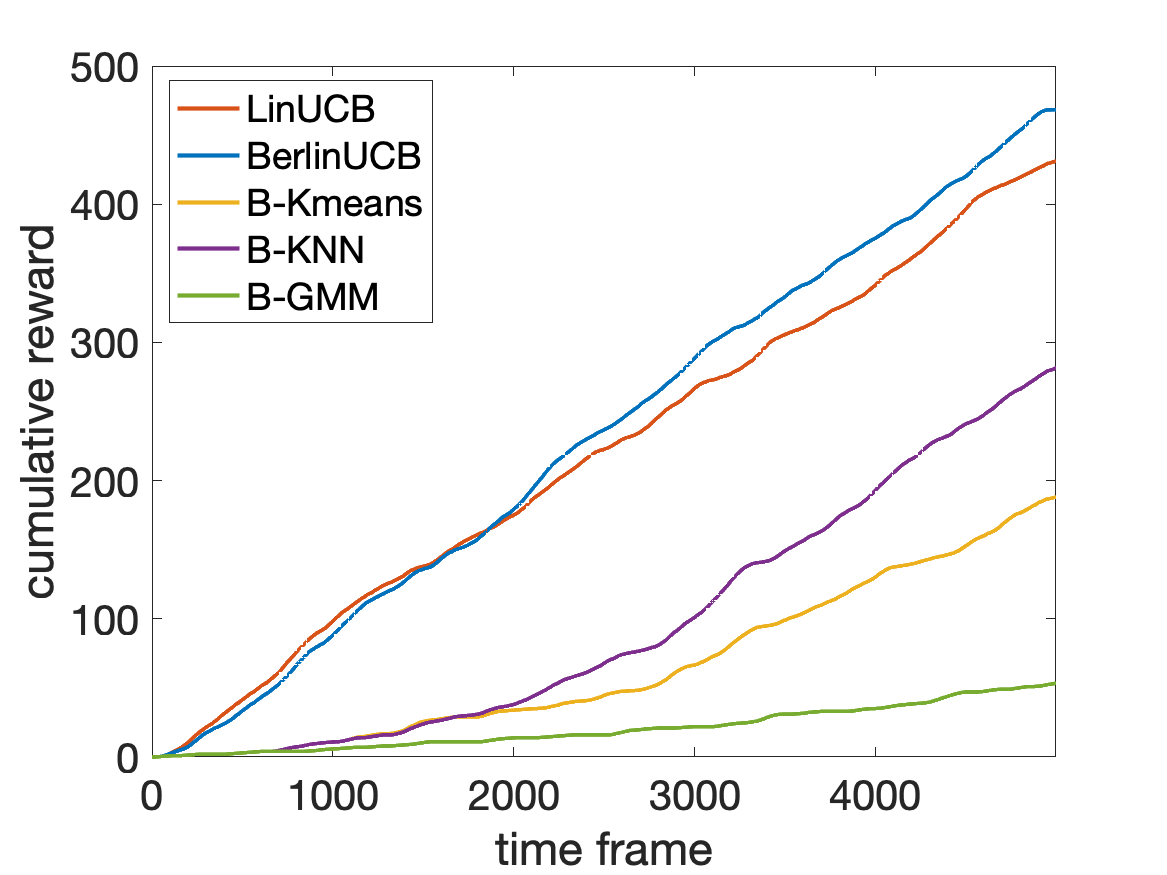} \hfill
    \includegraphics[width=0.24\linewidth]{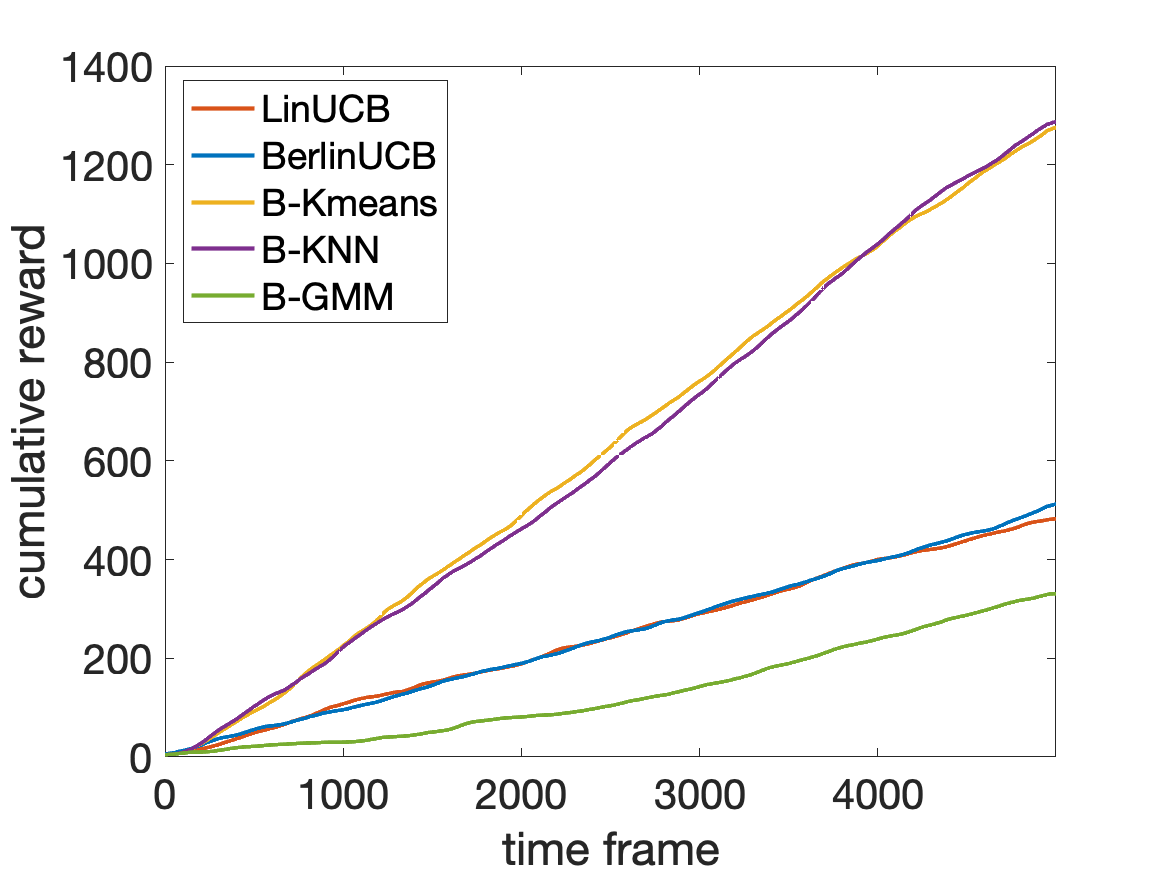} \hfill
    \includegraphics[width=0.24\linewidth]{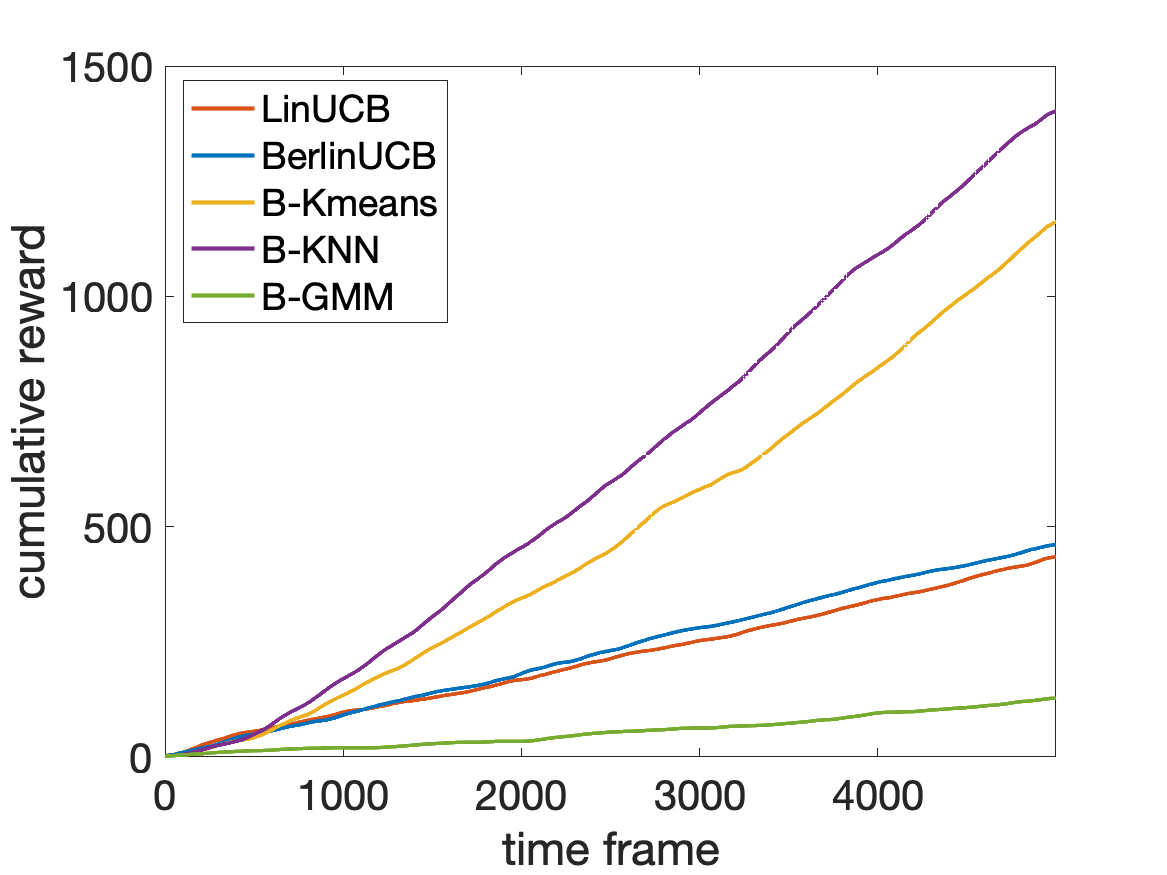}
    
\vspace{-0.2cm}
\par\caption{\textbf{Performance in MNIST}. Fixed arms: (a) stationary, $p_r$=0.01; (b) varying clusters, $p_r$=0.01; (c) negative images, $p_r$=0.1; (d) shuffled rewards, $p_r$=0.01. Extendable arms: (e) stationary, varying $p_r$; (f) varying clusters, $p_r$=0.01; (g) shuffled rewards, varying $p_r$; (h) shuffled rewards, $p_r$=0.1.}\label{fig:scores}
\end{figure}

\begin{table}[tbh]
\begin{minipage}{\linewidth}
      \caption{\textbf{Accuracy:} Stationary contexts with different probabilities of reward revealing}
      \label{tab:stat} 
      \centering
      \resizebox{\linewidth}{!}{
 \begin{tabular}{ l | c | c | c | c | c | c  | c | c }
 &\multicolumn{2}{c}{MNIST (varying $p_r$)} \vline & \multicolumn{2}{c}{MNIST ($p_r$=0.5)} \vline & \multicolumn{2}{c}{MNIST ($p_r$=0.1)} \vline & \multicolumn{2}{c}{MNIST ($p_r$=0.01)}   \\
  & fixed arms & extendable  & fixed arms & extendable  & fixed arms & extendable & fixed arms & extendable  \\ \hline
LinUCB & 0.1134 & 0.094 & 0.1080 & 0.0962 & 0.0842 & 0.0762  & 0.0252   &  0.0902   \\ \hline
BerlinUCB & 0.1138 & 0.0990 &  0.1102  &   0.0990 &   0.1016 &  0.0926   & 0.0788  & 0.0896      \\
B-Kmeans & 0.2594 & 0.2130 &  0.2674 &   \textbf{0.2678} &   \textbf{0.3132} &  \textbf{0.2760} &   \textbf{0.1400}   & 0.0828      \\
B-KNN & \textbf{0.2642} & \textbf{0.2398} &  \textbf{0.2722}  &    0.2642 &   0.2954 &  0.2622  &  0.1384     &  \textbf{0.0938}             \\
B-GMM & 0.0958 & 0.0768 &  0.1060  & 0.0728 & 0.0958 & 0.0320 & 0.1034  &  0.0120 \\ 
 \end{tabular}
}
 \end{minipage}
\end{table}

\begin{table}[tbh]
\begin{minipage}{\linewidth}
      \caption{\textbf{Accuracy:} Nonstationary cases with different probabilities of reward revealing}
      \label{tab:nostat} 
      \centering
      \resizebox{\linewidth}{!}{
 \begin{tabular}{ l | p{1.5cm} | p{1.5cm} | p{1.5cm} | p{1.5cm} | p{1.5cm} |p{1.5cm} | p{1.5cm} | p{1.5cm} | p{1.5cm} | p{1.5cm} | p{1.5cm} | p{1.5cm} | p{1.5cm}}
  &\multicolumn{3}{c}{MNIST - varying clusters} \vline & \multicolumn{3}{c}{MNIST - negative images} \vline & \multicolumn{3}{c}{MNIST - shuffled rewards}  \\
 Fixed Arms & $p_r$=0.5 & $p_r$=0.1 & $p_r$=0.01 & $p_r$=0.5 & $p_r$=0.1 & $p_r$=0.01 & $p_r$=0.5 & $p_r$=0.1 & $p_r$=0.01 \\ \hline
LinUCB &   \textbf{0.1086}   & 0.0984  &  0.0690   & 0.1044   & 0.0920  & 0.0732  & 0.1172   & 0.0832  & 0.0512  
  \\ \hline
BerlinUCB &   0.1024   &   0.1002   &   0.0828    &   0.1016  &   0.0970  &  0.1016  &   0.1120    & 0.0966 & 0.0832      \\
B-Kmeans &    0.0952    &  \textbf{0.1096}    &  \textbf{0.1074}   &   0.1082     &    0.1036  &  0.0768 & \textbf{0.2776}  &    0.2878   &  \textbf{0.1618}     \\
B-KNN &    0.0984    &   0.1002      & 0.1050 &     \textbf{0.1762}    &    \textbf{0.1732}    & 0.1016 &    0.2600      &   \textbf{0.3162}   & 0.1556   \\
B-GMM &   0.1072  &   0.0974   &  0.1034 &   0.0954 &  0.1074  &  \textbf{0.1018} & 0.1158  &   0.1172 &  0.1062      \\ 
 \end{tabular}
}
 \end{minipage} 
 \begin{minipage}{\linewidth}
      \centering
      \resizebox{\linewidth}{!}{
 \begin{tabular}{ l | p{1.5cm} | p{1.5cm} | p{1.5cm} | p{1.5cm} | p{1.5cm} |p{1.5cm} | p{1.5cm} | p{1.5cm} | p{1.5cm} | p{1.5cm} | p{1.5cm} | p{1.5cm} | p{1.5cm}}
  &\multicolumn{3}{c}{MNIST - varying clusters} \vline & \multicolumn{3}{c}{MNIST - negative images} \vline & \multicolumn{3}{c}{MNIST - shuffled rewards}  \\
Extendable Arms  & $p_r$=0.5 & $p_r$=0.1 & $p_r$=0.01 & $p_r$=0.5 & $p_r$=0.1 & $p_r$=0.01 & $p_r$=0.5 & $p_r$=0.1 & $p_r$=0.01 \\ \hline
LinUCB &  0.0994   &  0.0930   &  0.0866  &  0.0888   & 0.0918   & \textbf{0.1014} & 0.1010   & 0.0872  &  0.1014  \\ \hline
BerlinUCB &    0.0930     &   0.0946    &    \textbf{0.0938}   &   0.0924    &    0.0930 &    0.0910    & 0.0994   &  0.0926 &   0.0918     \\
B-Kmeans & \textbf{0.1010}     &   \textbf{0.0990}     &     0.0380   &   0.0910    &  0.0712   &   0.0308   &     0.2478   &  0.2336   &  0.0654       \\
B-KNN &   0.0958   &  0.0970      &   0.0570  & \textbf{0.1780}     &   \textbf{0.1228}   &  0.0460    &     \textbf{0.2606}    & \textbf{0.2818}  &  \textbf{0.1016}       \\
B-GMM &    0.0756    &  0.0296  &   0.0108  &  0.0680 & 0.0338  &   0.0118 &  0.0720  &    0.0258 &   0.0180  \\ 
 \end{tabular}
}
 \end{minipage} 
 \vspace{-1em}
\end{table}

\begin{table}[tbh]
 \vspace{-1em}
\begin{minipage}{\linewidth}
      \caption{\textbf{Accuracy:} Nonstationary contexts with varying episodic rewards}
      \label{tab:nscont} 
      \centering
      \resizebox{\linewidth}{!}{
 \begin{tabular}{ l | c | c | c | c | c | c | c | c | c}
 &\multicolumn{4}{c}{varying cluster distribution} \vline & \multicolumn{4}{c}{negative images} \vline & average  \\
  & MNIST-F & MNIST-E & Warfarin-F & Warfarin-E & MNIST-F & MNIST-E & Warfarin-F & Warfarin-E \\ \hline
LinUCB & \textbf{0.1016} & 0.0946 & 0.4580 & \textbf{0.4646} & 0.0956 &  0.0912 & \textbf{0.4440} &  \textbf{0.4374}  & 0.2734  \\ \hline
BerlinUCB &  0.0986  & \textbf{0.0948} & 0.4814 &  0.4480 &   0.0988  & 0.0926 &   0.3980 &  0.3898  & 0.2628  \\
B-Kmeans & 0.1012  &  0.0944 & 0.3964 &   0.2898 &   0.1048 & 0.0854 & 0.3200 &   0.1758  & 0.1960   \\
B-KNN & 0.1012 &  0.0946 &  0.4638 &   0.4134 &   \textbf{0.1626} &  \textbf{0.1576} & 0.4188 &   0.3998   & \textbf{0.2765}  \\
B-GMM & 0.1010 &  0.0684 &  \textbf{0.5494} & 0.2208 & 0.0926 & 0.0694 &  0.3992 & 0.2002 & 0.2126  \\ 
 \end{tabular}
}
 \end{minipage}
 
\end{table}

\begin{table}[tbh]
\begin{minipage}{\linewidth}
      \caption{\textbf{Accuracy:} Nonstationary rewards with varying episodic rewards}
      \label{tab:nsrewd} 
      \centering
      \resizebox{\linewidth}{!}{
 \begin{tabular}{ l | c | c | c | c | c | c | c }
 &\multicolumn{4}{c}{shuffled class labels} \vline & \multicolumn{2}{c}{multi-task setting} \vline & average \\
  & MNIST-F & MNIST-E & Warfarin-F & Warfarin-E & MNIST/Warfarin-F & MNIST/Warfarin-E &  \\ \hline
LinUCB &   0.1080 & 0.0974 & \textbf{0.6464} & \textbf{0.6348} & 0.3496 & 0.3442 & 0.3634   
    \\ \hline
BerlinUCB &   0.1126 &    0.1036 &    0.6116 &  0.6080 & 0.3136 &  0.3071 & 0.3428
 \\
B-Kmeans &  0.2376  &    0.2566  &   0.5262  &   0.5152 &  0.3801 & 0.3690  & 0.3808    \\
B-KNN &    \textbf{0.2574} &  \textbf{0.2582} &  0.6278 &   0.6026  & \textbf{0.3833}  &   \textbf{0.4041}  & \textbf{0.4222}          \\
B-GMM &    0.0958  &  0.0664  & 0.5488 &   0.4038 &   0.2052 & 0.2375 & 0.2596 \\ 
 \end{tabular}
}
 \end{minipage}
  \vspace{-1em}
\end{table}

We explored different combinations of the above nonstationarities. Table \ref{tab:stat} summarizes our results for the setting where the contexts are stationary, we observe that BerlinUCB algorithms consistently outperforming the baseline with a large marginal ($> 20\%$ accuracy comparing to LinUCB's around $10\%$), ranked top in all 8 scenarios with different reward revealing probablity $p_r$. 

Table \ref{tab:nostat} summarizes our results for the nonstationary context due to \textit{varying cluster distribution}, data corruption with \textit{negative images}, and \textit{shuffled reward function}. As we can see, across different reward revealing probabilities $p_r$, BerlinUCB algorithms outperform the baseline on 16 out of 18 scenarios. 

If we take a closer look at the nonstationary context scenarios in Table \ref{tab:nscont} where the rewards are episodically revealed with varying $p_r$ across batches, we noticed that the baseline is performing fairly well in three scenarios in Warfarin dataset. However, if we consider the mean accuracy in the entire set of this experiment, the BerlinUCB algorithms has the highest average accuracy, suggesting the advantage of using unlabelled data or missing rewards. Moreover, if we take a look at the whole iteration history, for example, for MNIST dataset (Figure \ref{fig:scores} (a)), we observe that initially, the baseline CB (orange line) is considerably worse than semi-supervised approaches, and requires a larger number of iteration to finally catch up with them. Other examples in Figure \ref{fig:scores} also demonstrated different scenarios where certain self-supervision modules (KNN and Kmeans) are more helpful than the other ones, but in most cases BerlinUCB, even without clustering modules, outperforms LinUCB. 

Next, Table \ref{tab:nsrewd} summarizes our results in nonstationary reward settings with \textit{shuffled reward} function and mixed domain (\textit{multi-task}) setting.
Based on the mean accuracy in the entire experiment, the top two algorithms were: B-KNN (mean accuracy  42.22$\%$) and B-Kmeans (mean accuracy 38.08$\%$). Note that, with nonstationary (shuffled) labels, the reward accumulated by the baseline CB usually remains significantly below the reward of the best semi-supervised approaches, at all iterations (Figure \ref{fig:scores}). Thus, in a more challenging setting with both context and reward nonstationarities, the semi-supervised approaches can effectively improve upon the standard contextual bandit.

We also observed an interesting phenomenon where decreasing feedback actually increases the performance in some cases (as in Table \ref{tab:nostat}). This leads to open questions such as: What are the consequences if some feedbacks are not reliable? Can a threshold be given for the acceptable number of inaccurate feedbacks to achieve the same satisfactory level of performance? Does this behavior also exist in scenarios with non-binary reward structures? Ongoing directions include extending BerlinUCB's scalability across a series of situations in order to better understand its potential.

\section{Application: Interactive Speaker Recognition}

One important application of our online learning problem is the cold-start speaker recognition task, where new users can join a conversation at any time, and the interactive system should learn to recognize the voice profile of the newly joined user given a limited number of user feedbacks. Let's consider such an interactive speaker recognition system: At the start, there are only two buttons available:  ``No Speaker'' and ``New Speaker''. The agent chooses an arm by setting it to be highlighted. If it is correct, the user does not have to change it (unless it's ``New Speaker'', where the user needs to click on it to confirm creating a new arm). If incorrect, the user clicks on the right arm as a feedback. 

\textbf{Application background.} Unlike synthetic online learning dataset, where a bandit feedback is always given, here in the real-life interactive systems, the user might not give the feedback timely and constantly. Note that we assume that when the agent correctly identifies the speaker (or no speaker), the user (as the feedback dispenser) should send no feedbacks to the system by doing nothing. This brings an additional layer of complexity to this application, as the agent needs to learn from a negative feedback (no feedback), assuming that it is sometimes due to a lack of timely feedback (reward not revealed) and other times due to the user's approval (reward postively revealed). In another word, in an ideal scenario when the agent does a perfect job by correctly identifying the speaker all the time, we are not necessary to be around to correct it anymore (i.e. truly feedback free). As we pointed out earlier, this could be a challenge earlier on, because other than implicitly approving the agent's choice, receiving no feedbacks could also mean the feedbacks are not revealed properly (e.g. the human oracle took a break). To adapt BerlinUCB to this application, we first define our actions: an arm ``New'' to denote that a new speaker is detected, an arm ``No Speaker'' to denote that no one is speaking, and $N$ different arms ``User n'' to denote user $n$ speaking. 

\textbf{Synthetic benchmark.}
We evaluated the validity of our algorithm in a synthetic online learning speaker recognition task. MiniVox \cite{lin2020speaker} is an automatic framework to transform any speaker-labelled dataset into continuous speech datastream with episodically revealed label feedbacks. This benchmark is a light-weighted environment specifically designed to simulate two real-world challenges in interactive systems: (1) the speech data in real life is never truncated into pieces for an intelligent system to classify, and (2) the reward feedbacks (i.e. telling an intelligent system that it is incorrect) is never as timely and complete. MiniVox utilizes the speaker recognition dataset like VoxCeleb \cite{Nagrani19} to create randomly generated multi-speaker ``conversations'' in continuous streams. As a proof of concept, we randomly selected 10 speaker profiles and generated a ``multi-speaker conversation'' recordings of 60000 time frames. As the industrial standard, for each time frame, we extracted the Mel Frequency Cepstral Coefficients (MFCC) \cite{hasan2004speaker} in a sliding window fashion given the real-time audio input, and it will be served as the context for our contextual bandit agents. Here we assume the number of speakers are known, and  the reward streams are sparsified given a revealing probability of 0.1, 0.01 and 0.001 to simulate different levels of users' attendance in this interactive system.

\begin{figure}[tb]
\centering
    \includegraphics[width=0.32\linewidth]{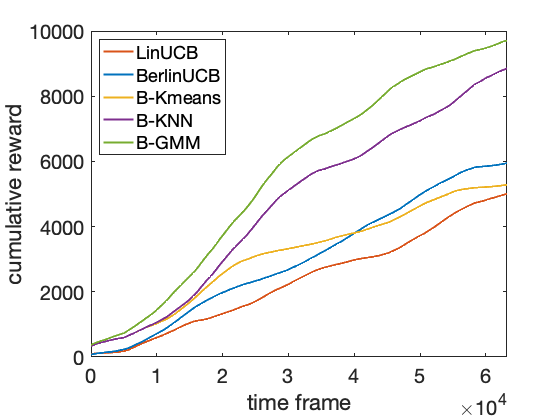} \hfill
    \includegraphics[width=0.32\linewidth]{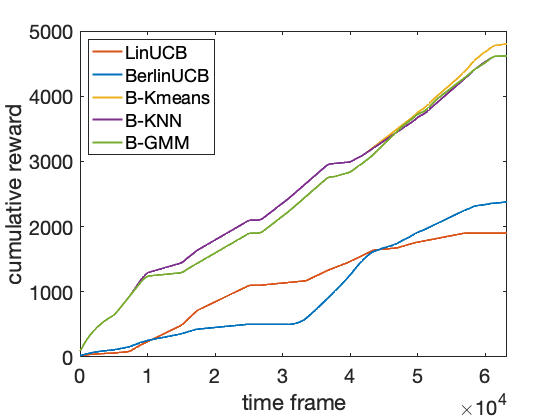} \hfill
    \includegraphics[width=0.32\linewidth]{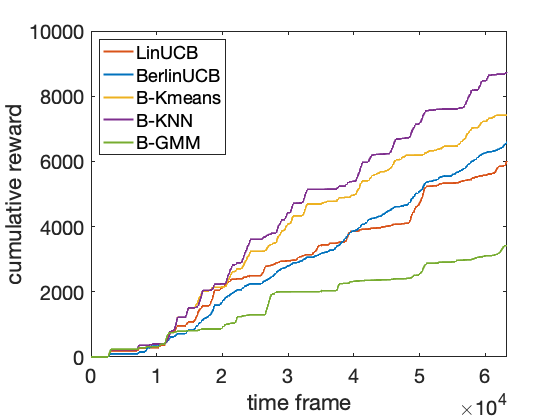}
\par\caption{\textbf{Cumulative rewards in MiniVox}. (a) $p_r$=0.1; (b) $p_r$=0.01; (c) $p_r$=0.001.}\label{fig:results}
\end{figure}

\textbf{Results.} As the common metric in online learning literature, we recorded the cumulative reward: at each frame, if the agent correctly predicts a given speaker, the reward is counted as +1 (regardless of whether the agent observes the reward or not). As shown in Figure \ref{fig:results}, BerlinUCB and its variants with clustering modules consistently outperforms LinUCB in the online speaker recognition task.


This application (as demostrated in \cite{lin2020voiceid}) provides an intriguing example of how our online learning algorithm can enable an interactive system to learn to recognize speaker identity (1) entirely escaping the necessity of registering user voiceprint beforehands, (2) effortlessly incorporating new users under an optimal exploration-exploitation trade-off, (3) effectively transferring representation of registered user features to new users, and (4) continually learning despite minimal involvement of human corrections (i.e. sparse and episodic feedbacks). 
\section{Conclusion}

We introduced an extension of the contextual bandit problem, learning from episodically revealed reward, motivated by several real-world applications in non-stationary environments, including recommendation systems, health monitoring and medical diagnosis, and others. In this setting, which we refer to as Contextual Bandit with Episodic Reward, the history of labeled and unlabeled contexts is available during online decision making, which allows the agent to learn context representations and reward mapping even when the reward is unobserved. We consider this problem an online semi-supervised learning problem, and take advantage of the clustering methods as a self-supervision modules to provide pseudo-reward feedback. The algorithms are evaluated in several types of nonstationary environments and compared to the standard  contextual bandit on several datasets. Overall, we observe clear advantages of the semi-supervised approaches over the standard contextual bandit; moreover, we demonstrated with a real-world application that the proposed algorithm can be beneficial for developing the interactive system that learns on the fly.

\bibliography{main}
\bibliographystyle{unsrt}

\end{document}